\documentclass[manuscript]{acmart}

\usepackage{amsmath,amsfonts}
\usepackage{algorithmic}
\usepackage{algorithm}
\usepackage{array}
\usepackage{textcomp}
\usepackage{stfloats}
\usepackage{multirow}
\usepackage{url}
\usepackage{verbatim}
\usepackage{graphicx}
\usepackage{booktabs}
\usepackage{graphicx}
\usepackage{subfigure}
\usepackage{cleveref} 
\usepackage{amssymb}
\usepackage{soul}

\AtBeginDocument{%
  }

\setcopyright{acmlicensed}
\copyrightyear{2024}
\acmYear{2024}
\acmDOI{XXXXXXX.XXXXXXX}

\begin{document}

\title{Neuro-Symbolic Embedding for Short and Effective Feature Selection via Autoregressive Generation}

\author{Nanxu Gong}
\authornote{Both authors contributed equally to this research.}
\orcid{https://orcid.org/0009-0006-4534-8395}
\affiliation{%
  \department{School of Computing and Augmented Intelligence}
  \institution{Arizona State University}
  \city{Tempe}
  \country{USA}
}
\email{nanxugong@outlook.com}

\author{Wangyang Ying}
\orcid{https://orcid.org/0009-0009-6196-0287}
\authornotemark[1]
\affiliation{%
  \department{School of Computing and Augmented Intelligence}
  \institution{Arizona State University}
  \city{Tempe}
  \country{USA}
}
\email{wangyang.ying@asu.edu}

\author{Dongjie Wang}
\orcid{https://orcid.org/0000-0003-3948-0059}
\authornote{Corresponding authors}
\affiliation{%
  \department{Department of Computer Science}
  \institution{University of Kansas}
  \city{Lawrence}
  \country{USA}}
\email{wangdongjie@ku.edu}

\author{Yanjie Fu}
\orcid{https://orcid.org/0000-0002-1767-8024}
\authornotemark[2]
\affiliation{%
  \department{School of Computing and Augmented Intelligence}
  \institution{Arizona State University}
  \city{Tempe}
  \country{USA}
}
\email{yanjie.fu@asu.edu}

\begin{abstract}
    Feature selection aims to identify the optimal feature subset for enhancing downstream models.
    Effective feature selection can 
    remove redundant features, save computational resources, accelerate the model learning process, and improve the model overall performance.
    However, existing works are often time-intensive to identify the effective feature subset within high-dimensional feature spaces.
    Meanwhile, these methods mainly utilize a single downstream task performance as the selection criterion, leading to the selected subsets that are not only redundant but also lack generalizability.
    To bridge these gaps, we reformulate feature selection through a neuro-symbolic lens and introduce a novel generative framework aimed at identifying short and effective feature subsets.
    More specifically, we found that feature ID tokens of the selected subset can be formulated as symbols to reflect the intricate correlations among features. 
    Thus, in this framework, we first create a data collector to automatically collect numerous feature selection samples consisting of feature ID tokens, model performance, and the measurement of feature subset redundancy.
    Building on the collected data, an encoder-decoder-evaluator learning paradigm is developed to preserve the intelligence of feature selection into a continuous embedding space for efficient search.
    Within the learned embedding space, we leverage a multi-gradient search algorithm to find more robust and generalized embeddings with the objective of improving model performance and reducing feature subset redundancy.
    These embeddings are then utilized to reconstruct the feature ID tokens for executing the final feature selection.
    Ultimately, comprehensive experiments and case studies are conducted to validate the effectiveness of the proposed framework.
    The associated data and code are publicly available~\footnote{\scriptsize\url{https://github.com/NanxuGong/feature-selection-via-autoregreesive-generation}}.
\end{abstract}


\renewcommand{\shortauthors}{Gong and Ying et al.}





\maketitle

\section{Introduction}

Feature selection refers to selecting the optimal feature subset that can greatly enhance downstream models.
Effective feature selection offers numerous benefits, such as the elimination of redundant features, conservation of computational resources, acceleration of the model learning and deployment, and enhancement of model performance.
This technique has been widely integrated into many research domains, such as bioinformatics, financial analysis, and urban computing, among others.

Existing works can be categorized into three categories: 1) Filter-based methods~\cite{yang1997comparative,forman2003extensive,hall1999feature,yu2003feature} identify the effective features that have the greatest predictive power according to the statistical measurement (\textit{e.g., Pearson Correlation, Chi-squared Value}); 2) Embedded methods \cite{tibshirani1996regression,sugumaran2007feature} derive feature importance for feature selection by integrating selection regularization with predictive loss during the optimization process; 3) Wrapper methods \cite{yang1998feature,kim2000feature,narendra1977branch,kohavi1997wrappers} formulate feature selection as a discrete searching problem and employ different optimization strategies (\textit{e.g., Evolutionary Algorithm, Reinforcement Learning}) to determine the best feature subset.

\begin{figure}[t]
    \centering
    \subfigure[Iterative selection view]{
    \begin{minipage}[ht]{0.4\linewidth}
    \centering
    \includegraphics[width=1.6in]{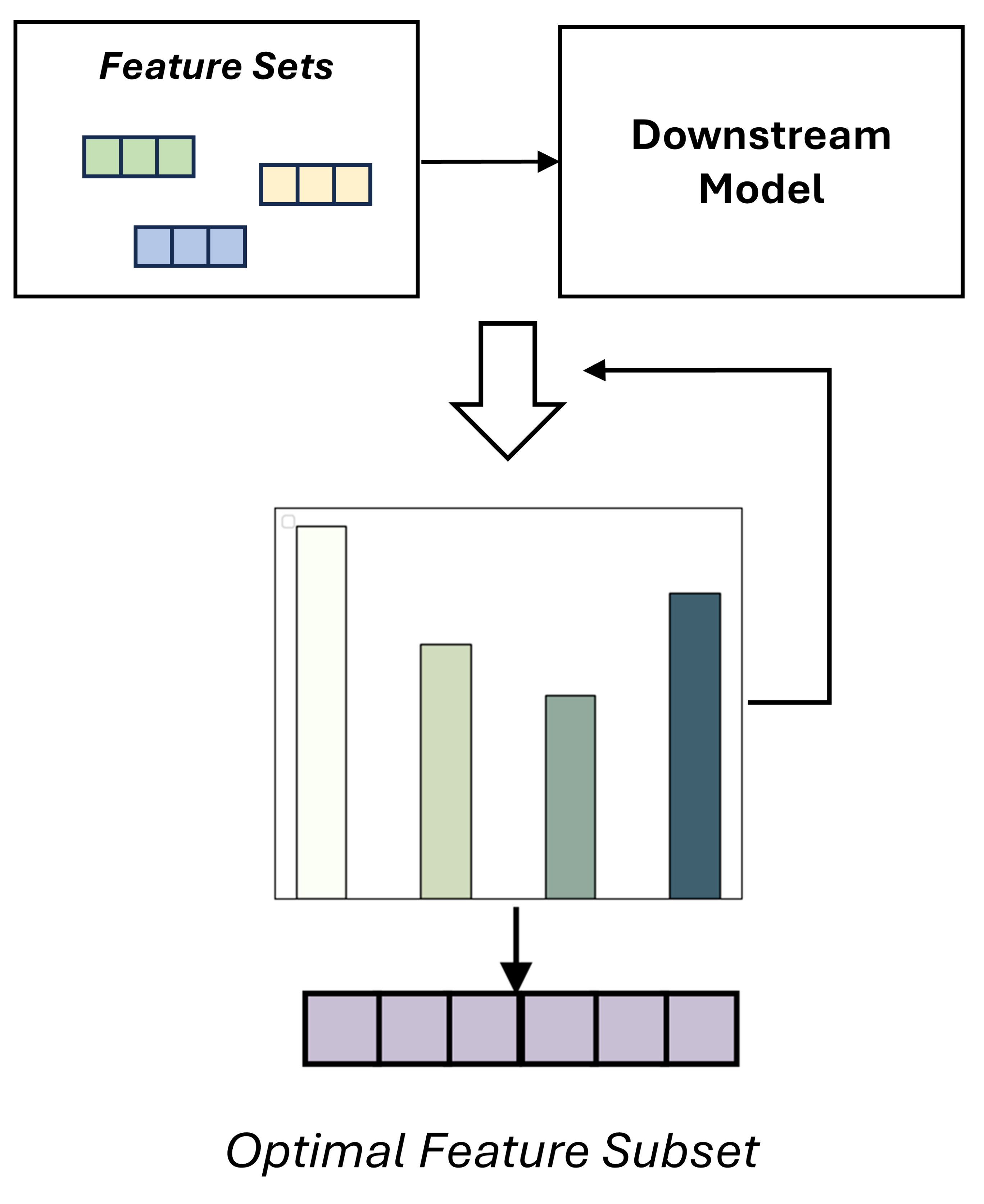}
    \label{method_contrast_1}
    \end{minipage}
    }
    \centering
    \subfigure[Generative view]{
    \begin{minipage}[ht]{0.4\linewidth}
    \centering
    \includegraphics[width=1.6in]{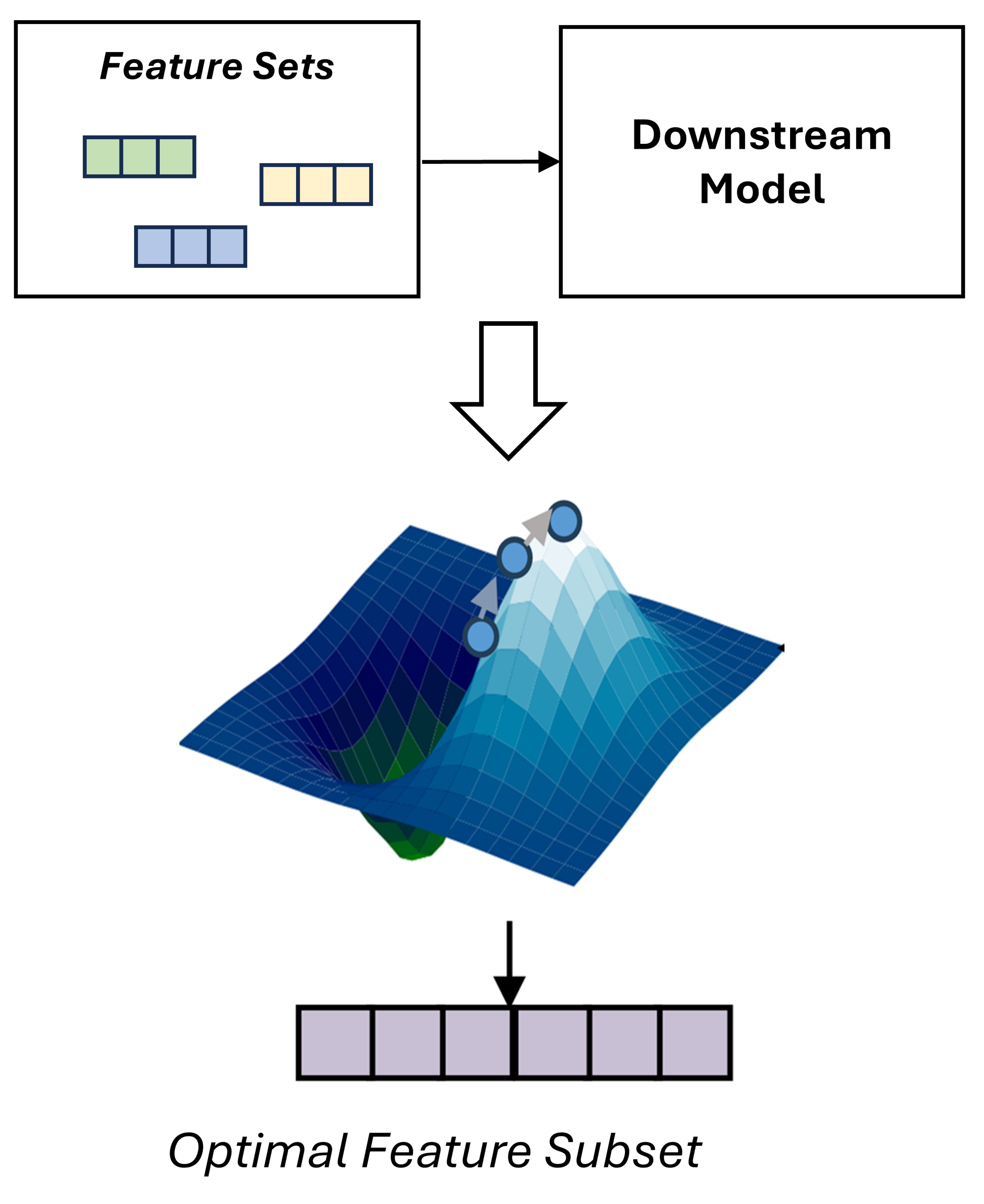}
    \label{method_contrast_2}
    \end{minipage}
    }
    \caption{We view feature selection from the generative perspective and reformulate it as a sequence generation task (b) rather than as an iterative discrete selection process (a).} \label{method_contrast}
    \vspace{-0.5cm}
\end{figure}

However, these approaches show various limitations in practical scenarios. Filter methods struggle to capture complex feature-feature correlations through simple statistical measurements test, resulting in suboptimal feature selection performance. Embedded methods, through their joint optimization process for predictive tasks and feature selection, strongly bind to specific downstream models, thus limiting their generalizability across different models. Wrapper methods, when applied to high-dimensional feature spaces, suffer from exponentially growing time complexity. Furthermore, most classical approaches predominantly optimize for single downstream task performance in feature selection, potentially leading to the selection  of redundant features vulnerable to overfitting the specific downstream task.

To overcome these limitations, it is significant to propose a novel feature selection framework, which can efficiently identify the short and effective feature subset.
However, there are two challenges: 
\begin{enumerate}
    \item \textbf{Challenge 1: Efficiently identifying effective features by considering complex feature-feature correlation.}
    Identifying the feature-feature correlations is crucial for selecting an effective feature subset. However, with the increase in feature space dimensionality, the time complexity associated with capturing these correlations for improved feature selection also rises correspondingly.

    \item \textbf{Challenge 2: Simultaneously considering both redundancy and effectiveness as  feature selection criteria. }
    Focusing solely on downstream task performance may cause a feature subset that is overly tailored to a specific model, making it not broadly applicable. However, balancing redundancy and effectiveness is challenging, as multi-objective search can introduce ambiguous search directions, potentially compromising performance.
    
\end{enumerate}

Upon analyzing the feature selection task, we found that treating the indicator of a feature as a symbol allows the pattern of these symbols to represent the knowledge inherent in feature selection. By analyzing the symbol patterns, we can assess the quality of the selected feature subset without directly involving the original data.
Thus, in this paper, we propose a novel \textbf{F}eature \textbf{S}election framework from a \textbf{N}euro-\textbf{S}ymbolic lens (\textbf{FSNS}), which can select the effective feature subset in an autoregressive manner.
In this framework, we initially establish a Reinforcement Learning (RL)-based data collection mechanism designed to autonomously collect a comprehensive array of feature selection samples. These samples consist of feature ID token symbols, corresponding model performance metrics, and an indicator for feature subset redundancy. Subsequently, leveraging the collected dataset, we construct an intricate encoder-decoder-evaluator learning framework. This architecture aims to preserve the intrinsic intelligence of feature selection processes within a continuous embedding space. Such preservation converts the discrete selection process into a continuous space optimization problem, thereby augmenting search efficiency and efficacy.
Furthermore, within the learned embedding space, we adopt a multi-gradient search algorithm to find more robust and generalized embeddings, with the dual objectives of enhancing downstream model performance and diminishing feature subset redundancy. These refined embeddings subsequently are used to reconstruct the ultimate feature selection symbols to execute the final feature selection process.
To demonstrate the effectiveness of our proposed framework, we undertook extensive experimental analyses and case studies across 16 different  real-world datasets.

\textbf{Our Contributions:}
\textit{1) Neuro-symbolic generative feature selection Learning:} We have redefined the feature selection process by transitioning from a traditional discrete decision-making model to a continuous optimization paradigm. This reformulation enhances the adaptability and efficiency of feature selection methods. 
\textit{2) Reinforcement learning-based data collector:} We introduce a novel RL-based data collection framework to autonomously gathers training data in both supervised and unsupervised settings, significantly reducing dependence on manual data collection.
\textit{3) Multi-objective optimization:} We incorporate both feature-feature redundancy and downstream task performance as dual objectives in the optimization process. This dual-objective approach uncovers succinct and effective feature subsets, alleviating the training burden and overfitting risk of downstream models, while yielding more robust and reliable results.

\section{Problem Statement}
We formulate the feature selection task as a sequential generative learning task. Given a dataset $D =\{X, Y\}$, where $X$ is a feature set and $Y$ is the target label. We collect the triplet group $R = \{\mathbf{f}_i, v_i, u_i\}^n_{i=1}$ as training data by applying feature selection method on $D$, where $\mathbf{f}_i = [f_1, f_2, ..., f_m]$ is a feature subset token sequence; $v_i$ is the performance; $u_i$ is the redundancy. 
In this paper, we preserve the knowledge of feature selection existing in $R$ into a continuous embedding space $\varepsilon$ by jointly optimizing an encoder $\phi$, a decoder $\psi$, a performance evaluator $\vartheta$, and a redundancy evaluator $\kappa$. 
We will search the best embedding $\mathbf{E}$ within the learned continuous embedding space and reconstruct the optimal feature subset token sequence $\mathbf{f}^*$ by:
\begin{equation}
    \mathbf{f}^* = \psi (\mathbf{e}^*) = argmax_{E\in \varepsilon} \mathcal{M}(X[\psi(\mathbf{E}), Y])
\end{equation}
where $\mathcal{M}$ is the downstream ML task. The feature subset token sequence $\mathbf{f}^*$ will be applied to the feature space $X$ to select the associated feature subset $X[\mathbf{f}^*]$ as the final output.

\section{Methodology}
\subsection{Framework Overview}
\begin{figure}[t]
    \centering
    \includegraphics[width=\linewidth]{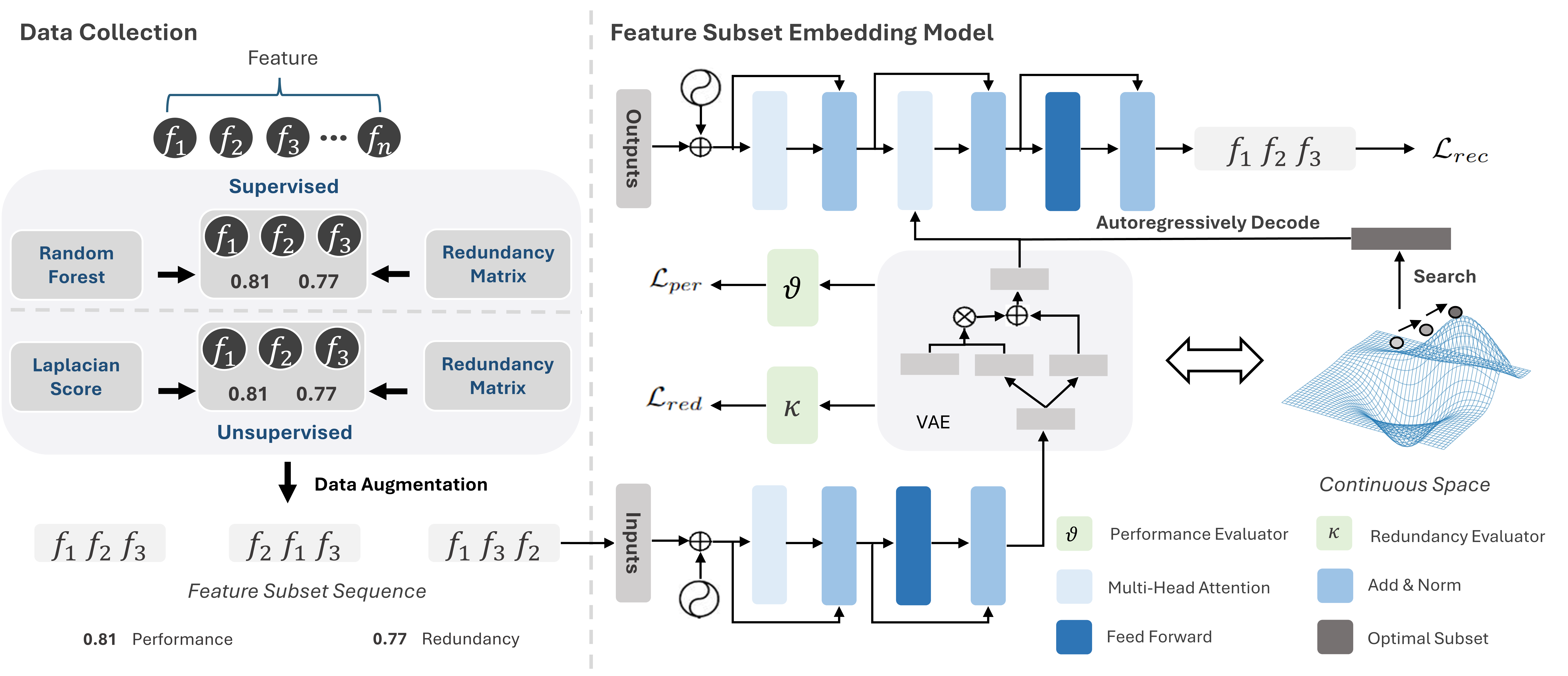}
    \caption{Framework overview. Firstly, we utilize the RL-based data collector to collect training data, consiting of feature subset sequence, model performance, and feature space redundancy. Then, we employ the feature subset embedding model to preserve feature learning knowledge into a continuous space and perform optimal subset search guided by gradient. }
    \vspace{-0.6cm}
    \label{fig:framework}
\end{figure}

Figure \ref{fig:framework} is the overview of the framework (\textbf{FSNS}), which includes two main components: 1) training data collection; 2) feature subset embedding model.
Specifically, the first component is used to collect training data, in which one sample is a triplet consisting of a feature subset (a set of feature ID token symbols), associated model performance and feature-feature redundancy. Thus, in the second component, we develop an encoder-decoder-evaluator framework for feature knowledge learning. Here, the encoder aims to convert feature subset sequences into a large differentiable continuous embedding space. Two evaluators are trained to estimate the performance and redundancy respectively given an embedding vector. The decoder reconstructs a feature subset sequence based on an embedding vector. Once the feature knowledge is preserved in the continuous space, we have converted the feature selection task as a gradient-driven optimization problem. 
Moreover, we search the best embedding along the gradient direction by maximizing model performance and minimizing feature-feature redundancy.
Finally, the identified embedding is utilized to reconstruct the corresponding feature subset sequence to get the optimal feature space.

\subsection{Data Collection}



Constructing an effective feature knowledge embedding space requires sufficient and diverse training data. However, collecting such data is time-consuming and labor-intensive. In addition, the quality of the training data will affect the search for the optimal feature subset. Therefore, we want the data collected to be both diverse and high-quality. To achieve this, we leverage reinforcement intelligence to build a reinforcement data collector to collect diverse, high-quality, and automated training data. We will introduce our process in two parts: performance collection and redundancy collection.

\noindent\textbf{Performance Collection.} Performance is employed to assess the utility of selected feature subsets. Our primary aim is to select the feature subsets that are beneficial for downstream tasks. The collection of performance can be achieved in both supervised and unsupervised manner.

\textit{1) Supervised Data Collection:}
Inspired by \cite{liu2019automating}, we developed an automated feature subset-performance exploration and collection system. In particular, to build a reinforcement data collector, we firstly create an agent for each feature. An agent can make selection decision for the corresponding feature. We regard the selected feature subset as a reinforcement learning environment and extract the statistical characteristics of the environment as a state, which can be achieved by various strategies (e.g., autoencoder). The reinforcement learning process involves many steps, each of which can be divided into two stages. i.e., control stage and training stage.

\ul{\emph{Control Stage:}} Each feature agent takes an action to select or deselect corresponding feature based on their policy networks. These actions will change the selected feature subset, leading a new environment. Also, an overall reward is generated through this process. To quantify it, we introduce a performance evaluator. Specifically, we adopt random forest as the performance evaluator. Because it is robust, stable, widely used in competitions, and can reduce the prediction accuracy variation caused by downstream models. For each feature subset, we can calculate the downstream task accuracy to most intuitively reflect its performance. In reward assignment, the reward is split equally and then assigned to each participating agent. Non-participating agents receive no reward.

\ul{\emph{Training Stage:}} For each agent, we employ a Deep Q-Network (DQN) \cite{mnih2015human} as the policy network. In the training stage, we leverage experience replay to train each of the policy networks independently. For agent $i$, at time $t$, we store a tuple \{$s^t_i, a^t_i, r^t_i, s^{t+1}_i$\} into its memory, where $s^t_i, a^t_i, r^t_i$, and $s^{t+1}_i$ denote the state, action, reward, and next state respectively. The networks are trained by corresponding samples to obtain the maximum long-term reward based on the Bellman Equation \cite{sutton1999reinforcement}:
\begin{equation}
    F(s^t_i,a^t_i|\theta^t) = r^t_i + \omega max  F(s^{t+1}_i, a^{t+1}_i|\theta^{t+1})
\end{equation}
where $\theta$ is the parameter of network $F$, and $\omega$ denotes the discount factor.

As reinforcement data collector continue to explore more feature subsets with better performance, we can balance both the diversity and quality of the feature subset-performance pairs to encode feature learning knowledge.

\textit{2) Unsupervised Data Collection:}
When we lack labeled data or when dealing with extremely large datasets which consumes a considerable amount of time and resources to conduct repeated downstream task prediction, it is difficult to collect sufficient and valid data in the supervised manner. To solve this problem, following the same framework, we further optimize the reinforcement data collector to deal with unsupervised data collection tasks. Since the performance of a feature set can be evaluated from multiple perspectives. In addition to downstream task accuracy, there are many other effective methods available in unsupervised condition. We propose a mean Laplacian Score method to quantify the performance of a feature subset by magnifying the individual-level importance to the set-level performance. Laplacian Score \cite{He2005laplacian} assigns a score to each feature in the dataset according to their locality preserving power, and finally take the $k$ features with the best performance as the final selected feature subset, which is a standard filter method. Since Laplacian Score can evaluate the importance of a feature, we can further leverage the mean Laplacian Score of a feature subset to represent its performance. Formally, let $x_i$ denotes $i$-th sample of dataset $X$, while $f_{ri}$ is the r-th feature value of $x_i$. For a feature $f_r$ which denotes $r$-th feature, its Laplacian Score can be computed by:
\begin{equation}
    L_r = \frac{\sum_{i,j}(f_{ri}-f_{rj})^2S_{ij}}{Var(f_r)}
\end{equation}
where $S_{ij}$ is the similarity between $x_i$ and $x_j$, while $Var(f_r)$ denotes the estimated variance of the $i$-th feature. A good feature is one that ensures two data points are in close proximity if and only if there exists an edge connecting these two points. In this case, the Laplacian Score tends to be small.

We firstly calculate the Laplacian Score of each feature in the raw dataset, forming a vector $V$. Since the smaller the Laplace Score, the more important a feature, we transform the vector by: $V = 1-V$. For a feature subset $s$, its performance can be quantified by:
\begin{equation}
    p_s = \frac{1}{n}\sum_{i=1}^{n}V_i
\end{equation}
where $n$ is the number of features in subset and $V_i$ is the Laplacian Score of $i$-th feature. 


\noindent\textbf{Redundancy Collection.}
The dataset often contains numerous redundant and irrelevant features that can adversely impact the accuracy of machine learning models and increase the risk of overfitting. One of the primary objectives of feature selection is to mitigate the redundancy inherent in feature sets and reduce the size of feature set. To emphasize this point, we also collect the corresponding redundancy for subsequent optimization.


Firstly, we formulate a redundancy matrix $R: N\times N$, where $N$ is the number of features in the raw dataset and $R_{ij}$ donates the relevance between $i$-th and $j$-th feature which can also be regarded as information redundancy. In our model, we employ three optional methods to quantify the redundancy among features: mutual information, covariance, and Pearson correlation coefficient.

1) \textit{Mutual Information} is often used to measure the degree of association between two random variables. Formally, given two variables $X$ and $Y$, the mutual information $ I(X;Y)$ is defined by:
\begin{equation}
    I(X;Y) = \sum_{x\in X} \sum_{y\in Y}{p(x,y)log(\frac{p(x,y)}{p(x)p(y))})}
\end{equation}
where $p(x,y)$ denotes the joint distribution of $X$ and $Y$, while $p(x)$ and $p(y)$ are corresponding marginal distribution. 

2) \textit{Covariance} can reflect the cooperative relationship between two variables and whether the change trend is consistent. The covariance between $X$ and $Y$ can be calculated by:
\begin{equation}
    Cov(X,Y) = [X-E(X)][Y-E(Y)]
\end{equation}
where $E(X)$ and $E(Y)$ are expectation of $X$ and $Y$ respectively. Since the covariance can be negative, we take absolute values for all the covariances we calculate.

3) \textit{Pearson Correlation Coefficient.} On the basis of covariance, the Pearson correlation coefficient eliminates the effect of different amplitude of variation of two variables, which is given by:
\begin{equation}
    \rho (X,Y) = \frac{Cov(X,Y)}{\sigma_X\sigma_Y}
\end{equation}
where, $\sigma_X$ and $\sigma_Y$ denotes the standard deviation of $X$ and $Y$. As with the covariance, we also take the absolute value of the calculated Pearson correlation coefficient.

Once the redundancy matrix is defined, we can quantify the redundancy of a feature subset $S$ by:
\begin{equation}
    r_s = \sum_{x_i, x_j \in S} R(x_i, x_j)
\end{equation}
where $R_{ij}$ is the redundancy of $i$-th and $j$-th feature.

\subsection{Data Augmentation}
Since our perspective is to treat feature subset embedding as deep sequential learning steered by encoder, decoder, and evaluators, it raises a challenge that the selected feature subsets are non-ordinal sets, making feature knowledge embedding unclear. In addition, data collection is a time-consuming process. Balancing the diversity of data and time is challenging. To solve this problem, we propose a shuffling-based augmentation method to represent a non-ordinal feature subset as a feature sequence. For instance, a triple data with an performance and redundancy, $\{(f_1, f_2, f_3), 0.83, 0.71\} $, can be converted into “$\{f_1f_2f_3, 0.83, 0.71\} $”. In classic computer vision research, people can use rotation or flipping to augment image data. Similarly, since a feature subset is order-agnostic, we can shuffle the feature sequence “$\{f_1f_2f_3, 0.83, 0.71\}$” to generate more semantically equivalent data points: “$\{f_2f_1f_3, 0.83, 0.71\}$”, “$\{f_3f_2f_1, 0.83, 0.71\}$” to augment training data. The shuffling augmentation can increase data sizes to fight data sparsity, improve data diversity, reduce data overfitting, construct an empirical training set that is closer to the true population, and improve embedding accuracy.

\subsection{Feature Subset Embedding Model}
In the discrete space, the feature knowledge and interactions between multiple features are difficult to learn. Following the spirit of chatGPT, we explore to embed discrete features into a continuous space, in order to better model and learn the feature knowledge.
To be specific, we propose an integrated encoder-decoder-evaluator framework, and its structure is illustrated in Figure \ref{fig:framework}. We employ the variational transformer to integrate not just the transformer \cite{vaswani2017attention} ’s foundational sequential embedding ability, but also the variational learning’s data distribution smoothing ability of Variational AutoEncoder \cite{kingma2013auto}.

\ul{\emph{The Encoder}} aims to embed a feature subset sequence into an embedding vector. we adopt a transformer ($\phi$) as the backbone leveraging self-attention mechanism to capture the potential correlations between sequences. Formally, consider a feature subset token sequence $\mathbf{f}_i = [f_1, f_2, ..., f_m]$.
We firstly map $\mathbf{f}_i$ into embedding vector. The input representation $\mathbf{v}_i = [v_1, v_2, ..., v_m]$ is formed by adding up the vector and its position embedding. Then the $\mathbf{v}_i$ is transformed to a query matrix $Q$, a key matrix $K$, and a value matrix $V$. The self-attention is applied to $Q, K,V$:
\begin{equation}
    Attention(Q, K, V) = Softmax(\frac{QK^T}{\sqrt{d_k}})V
\end{equation}
where $d_k$ is the dimension of $Q$ and $K$. The queries, keys, and values undergo linear projection $h$ times to enable the model to collectively focus on information across various representations, ultimately concatenating the results. A feed-forward network is applied to obtain the final embedding.
We then introduce a VAE-like component to smooth the feature subset embedding $\mathbf{e}^*$. In particular, our strategy is to add two fully connected layers behind the transformer to estimate the mean $\mathbf{m}$ and variance $\sigma$ of the distribution. After that, we sample a noised vector $\varepsilon$ from normal distribution. Then a embedding vector $\mathbf{e}^*$ can be sampled by: $\mathbf{e}^* = \mathbf{m} + \varepsilon * exp(\sigma)$.

\ul{\emph{The Decoder}} aims to reconstruct a feature subset token sequence. The neural architecture of the decoder is based on a Transformer. Subsequently, we add a softmax layer after the Transformer decoder to estimate the probability of the next feature token and infer the next token of the feature subset sequence in an autoregressive manner. Formally, assuming the feature token to decode is $f_d$, and the partially completed decoded feature subset token sequence is $f_1...f_{d-1}$. The probability of the $d$-th token is given by:
\begin{equation}
    P_{\psi}(f_d\vert \mathbf{e}^*, [f_1,f_2,...,f_{d-1}]) = \frac{exp(z_d)}{\sum^D_{c=1}exp(z_c)}
\end{equation}
, where $z_d$ represents the $d$-th output of the softmax layer, $\psi$ is the function notation of the decoder. The joint likelihood
of the entire feature subset token sequence under the decoder
is
\begin{equation}
    P_{\psi}(x^D\vert \mathbf{e}^*) =  \prod^D_{d=1} P_{\psi}(f_d\vert \mathbf{e}^*, [f_1,f_2,...,f_{d-1}])
\end{equation}

\ul{\emph{The Evaluator}} aims to evaluate the predictive performance and redundancy of a feature subset, thereafter, provide gradient information for fast searching of the optimal feature subset in the embedding space. 
We impose two evaluators after the encoder to evaluate the predictive performance and redundancy respectively. Both of them consist of a fully-connected neural layer, we define the performance evaluator as $\vartheta$ and the redundancy evaluator as $\kappa$. Therefore, the performance and redundancy of a given feature subset embedding $\mathbf{e}^*$ can be given by:
$\Ddot{v} = \vartheta (\mathbf{e}^*)$ and $\ddot{r} = \kappa (\mathbf{e}^*)$.

\ul{\emph{Joint Optimization.}} The encoder, decoder, and evaluator jointly steer the embedding of feature learning knowledge and learn a feature subset embedding space. The goal is to minimize the reconstruction loss and evaluator loss of feature subset embedding, as well as the KL divergence loss between the learned feature subset embedding space distribution and normal distribution. Specifically, we use the Mean Squared Error (MSE) to measure the evaluator loss in predicting feature subset predictive power:
\begin{equation}
    \mathcal{L}_{per} = MSE(v,\Ddot{v})
\end{equation}
Similarly, the redundancy evaluator loss is given by:
\begin{equation}
    \mathcal{L}_{red} = MSE(r,\ddot{r})
\end{equation}
We use the negative log-likelihood of generating feature subset sequential tokens to measure reconstruction loss:
\begin{equation}
\begin{split}
    \mathcal{L}_{rec} &= -logP_{\psi}(\mathbf{x}^{\mathbf{D}}\vert \mathbf{e}^*)\\
    &= -\sum^D_{d=1}logP_{\psi}(f_d\vert \mathbf{e}^*, [f_1,f_2,...,f_{d-1}])
\end{split}
\end{equation}
We use the KL divergence to measure the alignment loss between the normal distribution and the feature subset embedding space distribution:
\begin{equation}
    \mathcal{L}_{kl} = \sum (exp(\sigma_i) - (1 + \sigma_i) + (m_i)^2)
\end{equation}
Finally, we integrate the four losses to form a joint loss:
\begin{equation}
   \mathcal{L} = \alpha \mathcal{L}_{per} + \beta \mathcal{L}_{rec} + \gamma \mathcal{L}_{kl} + \delta \mathcal{L}_{red}
\end{equation}
$\alpha$, $\beta$, $\gamma$, and $\delta$ are the hyper-parameters to balance the influence of the three loss functions during the learning process.

Since VAE models have a famous posterior collapse problem, in which the KL divergence disappears and the model ignores the signal from the encoder. Most of the existing work using variational transformer relies on large-scale pre-training to mitigate the problem \cite{li2020optimus}. Inspired by \cite{park2021finetuning}, we choose a simple two-stage training style to avoid posterior collapse. The specific training steps are as follows: 1) \textit{\textbf{pre-training stage:}} We set $\gamma$ (the weight of KL loss) to 0 and leave the rest unchanged, similar to the training of an autoencoder; 2) \textit{\textbf{fine-tuning stage:}} $\gamma$ is recovered and the whole model is trained.

\subsection{Optimal Embedding Search and Reconstruction}
Generally, feature selection searches for an optimal feature subset in a discrete space. However, in high-dimensional datasets, such a search consumes a lot of time and does not cover all possible feature combinations. To avoid it, we perform optimization search on the feature subset embedding space using gradient-based search method.
After learning the embedding space of feature subsets, we can leverage the gradients from both the performance evaluator and the redundancy evaluator to iteratively find the optimal embedding. Specifically, assuming we pick a starting point $\mathbf{e}^*$ which is selected from the training data, we next update the target optimal embedding with a certain number of steps (e.g., $\eta$ steps) using a gradient ascent method. The directed gradient toward the best performance and minimum redundancy is inferred by exploiting the evaluators. Formally, the optimal feature subset embedding is given by:
\begin{equation}
    \mathbf{e}^+ = \mathbf{e}^* + \eta \{\frac{\partial [\vartheta(\mathbf{e}^*)]}{\partial \mathbf{e}^*} - \lambda \frac{\partial [\kappa(\mathbf{e}^*)]}{\partial \mathbf{e}^*}\} 
\end{equation}
where $\lambda$ is a trade-off parameter.

Once we identify the optimal feature subset embedding $\mathbf{e}^+$ through our gradient-ascent search, we can decode the optimal embedding to iteratively generate the underlying feature subset token sequence corresponding to the optimal embedding. Formally, in the $d$-iteration, assume the decoder already generates a partial feature subset token sequence: $f_1...f_{d-1}$. The feature token to generate is $f_d$. The probabilistic inference of $f_d$ under maximize likelihood estimation is
\begin{equation}
    f_d = argmax(P_{\psi}(f_d\vert \mathbf{e}^+, [f_1,f_2,...,f_{d-1}]))
\end{equation}
The decoder keeps generating the next feature token in an iterative and autoregressive fashion, until the decoder generates an end token (i.e., $EOS$), for instance, “$f_2f_6f_5EOS$”, representing the optimal feature subset: $\{f_2, f_5, f_6\}$.





\section{Experimental Results}
We conducted experiments on 19 datasets to answer the following questions: 
1) Can our neuro-symbolic feature selection learning method improve downstream tasks? 
2)  Is variational transformer-based configuration better than other encoder-decoder configurations? 
3) Is our unsupervised strategy effective and efficient? 
4) How does redundancy minimization influence the sizes of generated feature sets?
5) How does our RL data collector and data augmentation method influence the effectiveness of the model? 
6) Is our method robust when various machine learning models are used as a downstream task? 
7) How does the time and space complexity of our method change over an increasing feature size?
8) Is our model robust to noises?

\subsection{Experimental Setup}

\noindent\textbf{Data Description:}
We collected 19 datasets, each of which represents one application domain, from UCIrvine, and OpenML. 
We evaluated our method and baseline methods on two widely-used predictive tasks in machine learning: 1) classification (C); 2) regression (R). 
Table \ref{table:overall} shows the key statistics of the data sets, including predictive task type, number of instances, and number of features. 

\noindent\textbf{Evaluation Design:}
We aim to evaluate whether our AI-enabled neuro-symbolic feature selection method can improve a downstream predictive model. We used Random Forest as our downstream model. This is because Random Forest is an ensemble learning model widely used in many competitions. Random Forest is stable and robust. This allows us to control variations and uncertainty caused by a downstream predictive model, and better observe the performances of our feature selection learning model. 
We used the Relative Absolute Error (RAE) to measure the performance of a selected feature subset. A higher RAE score indicates a higher quality subset of features. 
For each of the 16 domain data sets, we randomly split the data set into two subsets (A and B).  
A was seen by our method. We used A to collect training data of selected features, performance, and redundancy(e.g. $\{(f_1, f_2, f_3), 0.83, 0.71\} $). 
In each training data, the selected features were regarded as a symbolic token sequence, along with the performance and redundancy score. 
Such symbolic token representations were regarded as feature selection knowledge base. Our generative model learned the knowledge about feature selection from such a neuro-symbolic format. 
B was not seen by our method. After the best feature subset was identified (e.g., $f_2f_5f_6$) by learning from A, we applied the identified feature subset to B in order to evaluate our selected features.

\noindent\textbf{Hyperparameters and Reproducibility:}

\noindent\underline{\textit{Data Collector:}} We used reinforcement learning agents to explore 300 epochs to collect all the possibilities of feature ID token symbols as training data and randomly shuffled each feature sequence 25 times to augment the training data.

\noindent\underline{\textit{Feature Subset Embedding: token sequences to vectors:}} 
We mapped feature tokens to a 64-dimensional embedding, and used a 2-layer network for both encoder and decoder, with a multi-head setting of 8 and a feed-forward layer dimension of 256. 
The latent dimensionality of the variational autoencoder was set to 64. 
The feature subset utility estimator consists of a 2-layer feed-forward network, in which  each layer was of 200 dimensions. The values of $\alpha$, $\beta$, and $\gamma$ were 0.8, 0.2, and 0.001, respectively. When we calculated redundancy, the values of $\alpha$, $\beta$, $\gamma$ and $\delta$ were 0.5, 0.3, 0.001 and 0.2. We set the batch size to 64 and the learning rate to 0.0001.

\noindent\underline{\textit{Optimal Embedding Search and Reconstruction:}} We used the top 25 feature sets to search for the feature subsets and kept the optimal feature subset. The value of $\lambda$ was set to 0.1.

\noindent{\textbf{Environmental Settings:}} 
All experiments are conducted on the Ubuntu 22.04.3 LTS operating system, Intel(R) Core(TM) i9-13900KF CPU@ 3GHz, and 1 way RTX 4090 and 32GB of RAM, with the framework of Python 3.11.4 and PyTorch 2.0.1. 

\noindent\textbf{Baseline Algorithms for Overall Comparisons}:
We compared our method with 11 widely used feature selection algorithms: 1) \textbf{STG} \cite{yamada2020feature} selects features based on probabilistic relaxation of the $\ell_0$ norm of features; 2) \textbf{RRA} \cite{seijo2017ensemble} gathers different feature sets and integrates them based on statistical sorting distributions. 3) \textbf{K-BEST} \cite{yang1997comparative} selects the top-k features with the highest importance scores; 4) \textbf{mRMR} \cite{peng2005feature} selects a feature subset by maximizing relevance with labels and minimizing feature-feature redundancy; 5) \textbf{LASSO} \cite{tibshirani1996regression} shrinks the coefficients of useless features to zero by sparsity regularization to select features; 6) \textbf{RFE} \cite{granitto2006recursive} recursively deletes the weakest features; 7) \textbf{GFS} \cite{fan2021autogfs} is a group-based feature selection method via interactive reinforcement learning; 8) \textbf{MCDM} \cite{hashemi2022ensemble} ensemble feature selection as a Multi-Criteria Decision-Making problem, which uses VIKOR sort algorithm to rank features based on the judgment of multiple feature selection methods; 9) \textbf{LASSONet} \cite{lemhadri2021lassonet} (LNet) is a neural network with sparsity to encourage the network to use only a subset of input features; 10) \textbf{SARLFS} \cite{liu2021efficient} is a simplified version of MARLFS to leverage a single agent to replace multiple agents to decide selection actions of all features; 11) \textbf{MARLFS} \cite{liu2019automating} uses reinforcement learning to create an agent for each feature to learn a policy to select or deselect the corresponding feature, and treat feature redundancy and downstream task performance as rewards.

\noindent\textbf{Variant Design of Our Framework for Internal Comparisons}:
To justify the impacts of different technical components of our method, we introduced several variants of the proposed model: 1) \textbf{FSNS$^*$} removes the variational inference component and solely uses the Transformer to create the feature subset embedding space; 2) \textbf{FSNS$^\prime$} adopts LSTM \cite{hochreiter1997long} to construct the feature subset embedding space; 3) \textbf{FSNS$^-$} collect training data in an unsupervised manner, using Laplacian Score to quantify the performance of feature subset; 4)\textbf{FSNS$^+$} is also an unsupervised variant model, but guided by the mean redundancy; 5) \textbf{FSNS$^{M, C, P}$} denotes the redundancy minimized model using mutual information, covariance, and Pearson correlation coefficient respectively.

\begin{table}[t]
    \centering
    \caption{Comparison of the results of the proposed method with seven widely used algorithms, with the best and second-best methods highlighted in \textbf{bold} and \underline{underline}. Additionally, we marked the performance improvement ratio achieved by our method.}
    \label{table:overall}
    \resizebox{\linewidth}{!}{
    \begin{tabular}{ccccccccccccccccc}
   \toprule\toprule
    Dataset &  Task & \#Sample &  \#Features &Original & STG & RRA & K-Best & mRMR & LASSO & RFE & GFS & MCDM & LNet & SARLFS & MARLFS & FSNS \\
    \midrule
    SpectF & C & 267 & 44 & 75.96 & 75.01 & 79.16  & \underline{81.41} & 79.16 & 75.96 & 80.80 & 75.96 & 79.16 & 79.16 & 78.21 & 75.01 & \textbf{89.78(+8.37\%)}\\
    SVMGuide3 & C & 1243 & 21 & 77.81 & 77.06 & 80.28 & 76.54 & 76.54 & 75.29 & 79.43 & 76.95 & 80.28 & 74.87 & \underline{82.48} & 78.32 & \textbf{84.11(+1.63\%)}\\
    German Credit & C & 1001 & 24 & 64.88 & 63.40 & 67.07 & 65.28 & 67.63 & 65.87 & 67.88 & 65.51 & 69.21 & 58.09 & \underline{69.24} & 68.02 & \textbf{72.83(+3.59\%)}\\
    UCI Credit & C & 30000 & 25 & 80.19 & 79.96  & 74.88 & \underline{80.48} & \underline{80.48} & 77.94 & 80.16 & 80.16 & 74.39 & 74.11 & 80.05 & 80.10 & \textbf{80.78(+0.30\%)}\\
    SpamBase & C & 4601 & 57 & 92.68 & 92.03 & 91.39 & 91.47 & 92.02 & 91.73 & \underline{92.13} & 91.84 & 90.93 & 89.08 & 91.28 & 91.72 & \textbf{93.23(+1.10\%)}\\
    Ap\_omentum\_ovary & C & 275 & 10936 & 66.19 & 82.80 & \underline{84.49} & 82.80 & \underline{84.49} & 82.03 & 83.19 & 83.02 & \underline{84.49} & 83.02 & 82.09 & \underline{84.49} & \textbf{86.52(+2.03\%)} \\
    Ionosphere & C & 351 & 34 & 92.85 & 88.35 & 85.93 & 91.38 & \underline{95.69} & 86.98 & 95.69 & 94.22 & 91.41 & 85.58 & 94.16 & 95.68 & \textbf{97.10(+1.41\%)}\\
    Activity & C & 10299 & 561 & 96.17 & 95.39 & 95.39 & 95.78 & 95.87 & 95.34 & \underline{96.17} & 96.12 & 95.97 & 94.47 & 95.78 & 95.83 & \textbf{96.60(+0.43\%)}\\
    Mice-Protein & C & 1080 & 77 & 74.99  & 76.42 & 74.53 & 80.11 & 80.54 & 78.27 & 79.14 & 76.85 & 74.04 & 74.99 & 74.10 & \underline{80.93} & \textbf{82.41(+1.48\%)}\\
    Amazon Employee & C & 32769 & 9 & 93.02 & 93.49 & 93.09 & \underline{93.50} & \underline{93.50} & 93.23 & 93.01 & 91.56 & 92.87 & 93.46 & 93.46 & 93.49 & \textbf{93.55(+0.05\%)}\\
    Higgs Boson & C & 50000 & 28 & 69.57 & 66.88 & 68.95 & \underline{69.90} & \underline{69.90} & 69.84 & 69.66 & 69.71 & 68.95 & 68.72 & 68.32 & 69.46 & \textbf{70.32(+0.42\%)}\\
    \midrule
    Openml-586 & R & 1000 & 25 & 54.95 & 13.17 & 58.67 & 57.68 & 57.29 & \underline{60.67} & 57.64 & 45.34 & 56.90 & 58.28 & 55.48 & 57.71 & \textbf{63.28(+2.61\%)}\\
    Openml-589 & R & 1000 & 25 & 50.95 & 49.59 & 52.02 & 54.09 & 54.03 & \underline{59.74} & 54.56 & \underline{59.74} & 52.96 & 54.30 & 55.07 & 53.46 & \textbf{61.13(+1.39\%)}\\
    Openml-607 & R & 1000 & 50 & 51.73 & 38.43 & 55.93 &  53.54 & 53.72 & \underline{58.10} & 55.14 & 41.94 & 54.93 & 54.25 & 57.11 & 55.41 & \textbf{60.89(+2.79\%)}\\
    Openml-616 & R & 500 & 50 & 15.63 & 16.57 & 23.83 & 22.69 & 23.44 & \underline{28.98} & 22.56 & 22.93 & 22.92 & 16.34 & 24.09 & 23.66 & \textbf{41.81(+12.83\%)}\\
    Openml-618 & R & 1000 & 50 & 46.89 & 3.84 & 45.36 & 51.79 & 51.79 & 47.41 & 51.08 & \underline{52.79} & 52.23 & 50.51 & 50.39 & 52.45 & \textbf{54.38(+1.93\%)} \\
    Openml-620 & R & 1000 & 25 & 51.01 & 32.29 & 54.74 & 57.48 & 57.20 & 57.99 & 58.01 & 45.17 & 54.92 & \underline{58.23} & 54.56 & 57.80 & \textbf{62.00(+3.77\%)}\\
    Openml-637 & R & 500 & 50 & 14.95 & -3.13 & 20.91 & 21.86 & 21.88 & \underline{26.02} & 15.10 & 4.26 & 23.41 & 3.90 & 19.45 & 17.10 & \textbf{39.49(+13.47\%)}\\
    Housing Boston & R & 506 & 13 & 42.13 & 41.87 & 42.09 & \underline{44.69} & 40.20 & 30.29 & 42.65 & 42.15 & 41.56 & 43.01 & 35.16 & 39.70 & \textbf{46.33(+1.64\%)} \\
    \bottomrule\bottomrule
    \end{tabular}}
\end{table}

\subsection{Experimental Results}

\subsubsection{A Study of Overall Comparisons}
This experiment aims to answer: \textit{Can the features selected by our method effectively improve downstream tasks compared with other baseline methods?}
We evaluated our method and nine baseline algorithms on 19 application datasets over three evaluation metrics. 
For each application domain data, the objective is to verify whether our method can identify better feature subsets from the data subset A and boost predictive performances on the data subset B. 
Table \ref{table:overall} shows our method generates better feature subsets than several widely-used baseline algorithms over all the 19 application domains, particularly outperforming the best baseline by an average of 3\%. 
Moreover, none of the baseline algorithms can consistently achieve the second-best performance across all domains. 
The performances of the baseline models shows the limitations of search based and thresholding based strategies, as well as specific model structural (sparsity) assumptions, making them incapable of generalizing to other domains. 
In contrast, our neuro-symbolic method exhibits better task-agnostic performance and robust performances over domains. 
This can be explained by our perspective of seeing feature selection as neuro-symbolic generative AI; that is, using a large deep sequential learning model to embed complex and mechanism unknown feature knowledge into a large embedding space and then formulate discrete feature selection into a deep generative AI problem under effective and automated continuous optimization.

\begin{figure*}[t]
    \centering
    \subfigure[UCI Credit]{
    \begin{minipage}[ht]{0.23\linewidth}
    \centering
    \includegraphics[width=1.3in]{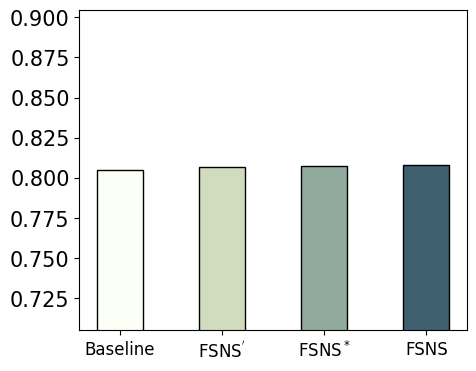}
    \end{minipage}
    }
    \centering
    \subfigure[SpamBase]{
    \begin{minipage}[ht]{0.23\linewidth}
    \centering
    \includegraphics[width=1.3in]{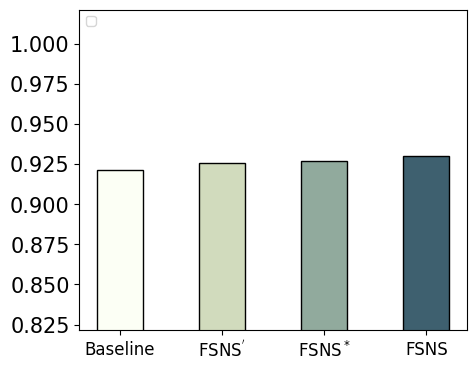}
    \end{minipage}
    }
    \subfigure[Activity]{
    \begin{minipage}[ht]{0.23\linewidth}
    \centering
    \includegraphics[width=1.3in]{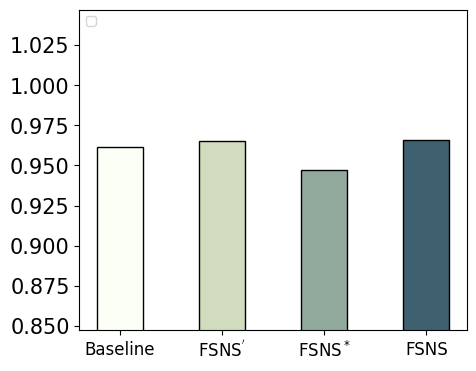}
    \end{minipage}
    }
    \subfigure[German Credit]{
    \begin{minipage}[ht]{0.23\linewidth}
    \centering
    \includegraphics[width=1.3in]{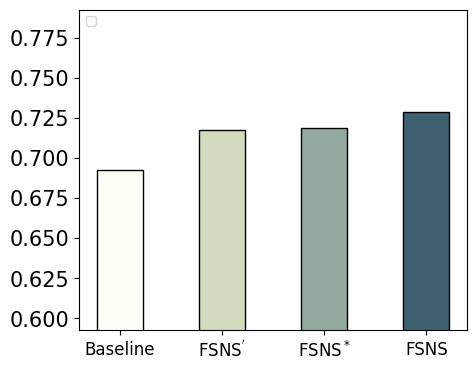}
    \end{minipage}
    }
    \subfigure[Openml-586]{
    \begin{minipage}[ht]{0.23\linewidth}
    \centering
    \includegraphics[width=1.3in]{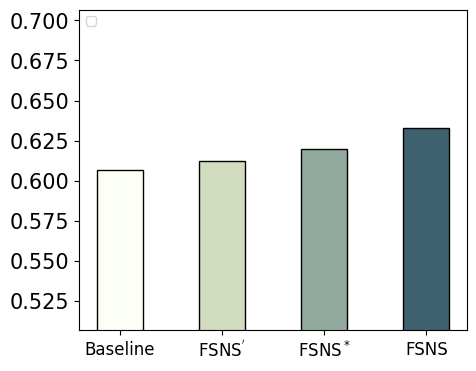}
    \end{minipage}
    }
    \subfigure[Openml-589]{
    \begin{minipage}[ht]{0.23\linewidth}
    \centering
    \includegraphics[width=1.3in]{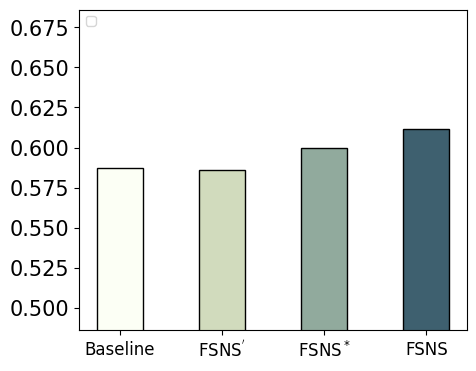}
    \end{minipage}
    }
    \subfigure[Openml-607]{
    \begin{minipage}[ht]{0.23\linewidth}
    \centering
    \includegraphics[width=1.3in]{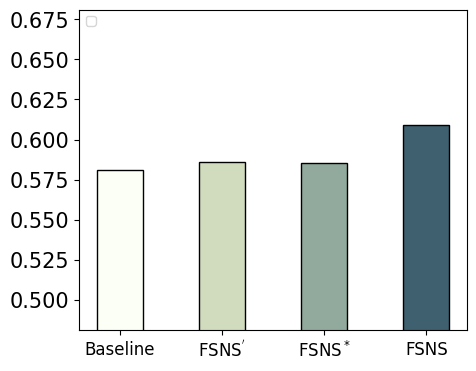}
    \end{minipage}
    }
    \subfigure[Openml-620]{
    \begin{minipage}[ht]{0.23\linewidth}
    \centering
    \includegraphics[width=1.3in]{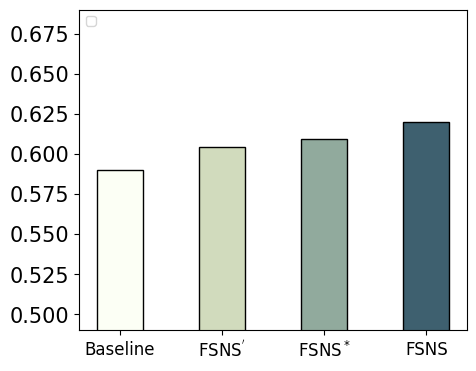}
    \end{minipage}
    }
    \caption{Results of different encoder-decoder configurations for downstream tasks, where FSNS$^\prime$ adopts LSTM as the backbone, FSNS$^*$ replaces LSTM with transformer, and FSNS combines VAE and transformer.} \label{fig:configuration}
    \vspace{-0.3cm}
\end{figure*}


\subsubsection{A Study of Integrating Variational Methods with Transformer}
This experiment aims to answer: \textit{why do we use the benefits of using variational transformer instead of other encoder-decoder configurations (e.g., LSTM, VAE, transformer)?} 
We evaluate our approach using three different encoder-decoder configurations. 
Initially, we employed LSTM as our sequential model, represented by FSNS$^{'}$, and Figure \ref{fig:configuration} demonstrates that our generative feature selection framework outperforms the baselines. 
Subsequently, we replaced LSTM with a transformer, denoted as FSNS$^*$ , and observed that FSNS$^*$ yielded better performance compared to FSNS$^{'}$. 
Finally, we compared our approach (denoted by FSNS) of integrating variational autoencoder into Transformer against a standalone transformer. 
Figure \ref{fig:configuration} shows the integration of variational learning into the Transformer consistently achieves higher accuracy.
This can be attributed to the fact that the variational transformer can more effectively estimate and smooth the feature subset embedding space, resulting in improved feature subset generation.

\subsubsection{On Effectiveness and Efficiency of Unsupervised Feature Selection}
\begin{figure}[t]
    \centering
    \subfigure[Downstream task accuracy]{
    \begin{minipage}[ht]{0.4\linewidth}
    \centering
    \includegraphics[width=1.6in]{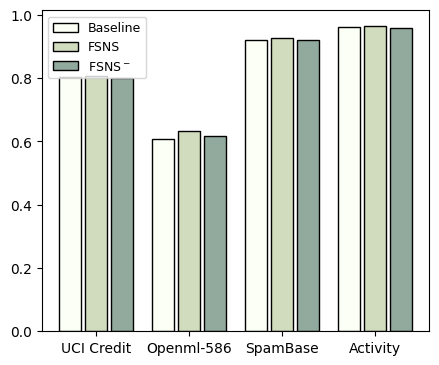}
    \label{exp:u1}
    \end{minipage}
    }
    \centering
    \subfigure[Time cost of RL data collector]{
    \begin{minipage}[ht]{0.4\linewidth}
    \centering
    \includegraphics[width=1.6in]{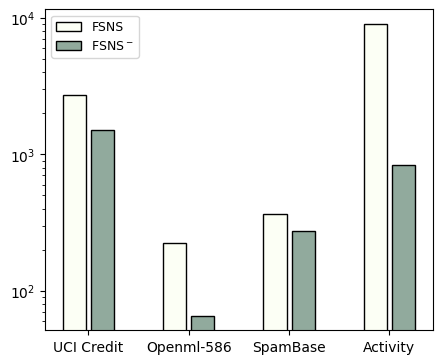}
    \label{exp:u2}
    \end{minipage}
    }
    \caption{Results of supervised and unsupervised data collection methods. We compared (a) downstream task accuracy across four datasets, and (b) RL data collection time.}
    \vspace{-0.3cm}
\end{figure}

This experiment aims to answer: \textit{Does our method perform well even without supervision?} We investigated the difference between unsupervised and supervised in terms of model performance. Figure \ref{exp:u1} shows that our proposed is a generic framework that can achieve excellent results even without supervision signals. We use FSNS$^-$ to denote the unsupervised version of our method. Feature selection guided by an unsupervised Laplacian score-based metric leads to an accuracy that is similar to using instead of supervised downstream task accuracy. The potential insight is that the Laplacian score can approximately evaluate feature importance, guiding the model to select significant features and eliminate redundant ones. This indicates that our model can be generalized to not just supervised setting but also unsupervised setting.
We further compared different utility metrics of unsupervised feature sets and examined their impacts on our framework. Specifically, FSNS$^+$ is our framework variant version that adopts feature set mean redundancy as a feature set utility metric. Table \ref{table:di_evaluation} shows that compared with redundancy, the aggregated Laplacian score-guided model can achieve better results on most of the datasets. When dealing with complex datasets, particularly noisy ones, the ability of redundancy as feedback to perform feature selection is rather limited. In contrast, the aggregated Laplacian score-guided model is more inclined to retain useful features while reducing the redundancy of the feature set.

Moreover, we study the time costs of training data collection when comparing unsupervised and supervised implementations. 
Figure \ref{exp:u2} shows that the unsupervised model can save a significant amount of time on training data collection, especially when dealing with large datasets. This is because using unsupervised strategies can avoid the need to repeatedly test downstream tasks to collect downstream task performance.

\subsubsection{On Effectiveness of Minimizing Redundancy}
This experiment aims to answer: \textit{How does the strategy of minimizing redundancy impact the quality and size of a selected feature subset?} 
Firstly, we studied the differences before and after applying the strategy of minimizing redundancy. Table \ref{table:redundancy} shows the results before and after redundancy minimization. 
Specifically, FSNS$^M$ denotes the redundancy minimized model using mutual information. Compared to FSNS, FSNS$^M$ can better reduce the redundancy of feature subsets with a minimal loss in accuracy. This suggests that the dual-objective optimization approach endows the model with the ability to discern succinct and effective feature subsets, thereby enhancing the application on downstream tasks with limited computational resources. We further examined the effect of using the number of features as feature redundancy, as shown in Table \ref{table:unsupervised}. Overall, all three methods are effective in reducing the number of features, especially in noisy datasets like openml-586. 
Another interesting finding is that by minimizing feature-feature redundancy, we can jointly improve feature subset performance and decrease feature sizes. This indicates that in the presence of noise within the dataset, reducing feature redundancy not only decreases the number of features but also eliminates noise, thereby enhancing the model's performance.

\begin{table}[t]
    \centering
    \caption{Results of using different performance evaluation methods (i.e., downstream model performance, Laplacian score, and feature set mean redundancy. )}
    \label{table:di_evaluation}
    \resizebox{65mm}{!}{
    \begin{tabular}{ccccc}
   \toprule\toprule
   Dataset & Original & FSNS & FSNS$^-$ & FSNS$^+$\\
    \midrule
    UCI Credit & 80.19 & 80.78 & 80.09 & 80.31\\     

    Activity & 96.17 & 96.60 & 95.73 & 91.52\\        

    Openml-586 & 54.95 & 63.28 & 61.13 & 19.77\\   
    Openml-589 & 50.95 & 61.13 & 59.76 & 39.36\\
    
    \bottomrule\bottomrule
    \end{tabular}}
\end{table}

\begin{table}[t]
    \centering
    \caption{Results before and after redundancy minimization. We compare downstream task accuracy and feature subset redundancy.}
    \label{table:redundancy}
    \resizebox{90mm}{!}{
    \begin{tabular}{cccccc}
   \toprule\toprule
   Dataset &  Performance & Original & Baseline & FSNS & FSNS$^M$\\
    \midrule
    \multirow{2}{*}{UCI Credit}& Accuracy & 80.19 & 80.48 (mRMR) & 80.78 & 80.40\\         
                           & Redundancy & 100 & 22.38 (mRMR) & 14.78 & 0.00 \\
                            \midrule
    \multirow{2}{*}{Activity}& Accuracy & 96.17 & 96.17 (RFE) & 96.60 & 93.88\\         
                           & Redundancy & 100 & 27.14 (RFE) & 23.36 & 0.76 \\
                           \midrule
    \multirow{2}{*}{Openml-586}& Accuracy & 54.95 & 60.67 (LASSO) & 63.28 & 62.73\\        
                           & Redundancy & 100 & 55.31 (LASSO) & 53.97 & 12.92 \\
                        \midrule
    \multirow{2}{*}{Openml-589} & Accuracy & 50.95 & 59.74 (LASSO) & 61.13 & 59.71 \\
                                & Redundancy & 100 & 35.26 (LASSO) & 34.55 & 18.91\\
    
    \bottomrule\bottomrule
    \end{tabular}}
\end{table}




\begin{table}[t]
    \centering
    \caption{Results of using mutual information, covariance, Pearson correlation coefficient as redundancy quantization method respectively.}
    \label{table:unsupervised}
    \resizebox{90mm}{!}{
    \begin{tabular}{ccccccc}
   \toprule\toprule
   Dataset &  Performance & Original & FSNS & FSNS$^M$ & FSNS$^C$ & FSNS$^P$ \\
    \midrule
    \multirow{2}{*}{UCI Credit}& Accuracy & 80.19 & 80.78 & 80.40 & 80.59 & 79.83\\         
                           & Number & 24 & 12 & 1 & 3 & 6\\
                            \midrule
    \multirow{2}{*}{Activity}& Accuracy & 96.17 & 96.60 & 93.88 & 94.08 & 95.58\\         
                           & Number & 561 & 255 & 43 & 88 & 65\\
                           \midrule
    \multirow{2}{*}{Openml-586}& Accuracy & 54.95 & 63.28 & 62.73 & 64.00 & 64.25\\         
                           & Number & 25 & 5 & 4 & 3 & 4\\
                           \midrule
    \multirow{2}{*}{Openml-589}& Accuracy & 50.95 & 61.13 & 59.71 & 61.13 & 59.76\\
                                & Number & 25 & 5 & 4 & 4 & 5 \\
    
    \bottomrule\bottomrule
    \end{tabular}}
\end{table}

\subsubsection{The Impacts of Reinforcement Data Collector and Data Augmentation.}
\begin{figure}[t]
    \centering
    \subfigure[Impact of RL data collector]{
    \begin{minipage}[ht]{0.4\linewidth}
    \centering
    \includegraphics[width=1.6in]{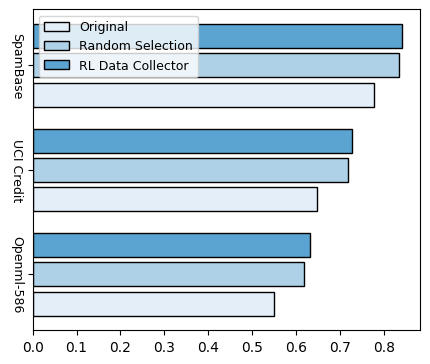}
    \label{exp:data_collection}
    \end{minipage}
}
    \centering
    \subfigure[Impact of data augmentation]{
    \begin{minipage}[ht]{0.4\linewidth}
    \centering
    \includegraphics[width=1.6in]{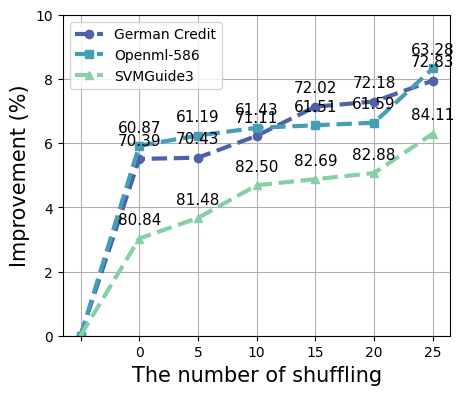}
    \label{exp:data_augmentation}
    \end{minipage}
}
    \caption{The effects of the RL data collector and data augmentation, (a) compare the downstream task accuracy achieved by random collection and the RL data collector, and (b) show the effect of different shuffling numbers on the accuracy}
    \vspace{-0.3cm}
\end{figure}

This experiment aims to answer: \textit{How does our RL data collector and data augmentation method impact model performance?} 
We conducted two control group experiments on three distinct datasets to investigate the roles of the data collector and data augmentation. 
Firstly, we compared the reinforcement learning-based data collector with random selection. Figure \ref{exp:data_collection} shows that random selection significantly decreased performance, even falling below baseline performance levels in several instances. This can be explained by the fact that the reinforcement learning-based data collector, in comparison to random selection, has the capacity to explore high-quality and diverse training data. Consequently, it constructs a more effective embedding space to advance the search of optimal feature subsets. 
Moreover, we compared the performances with data augmentation with the performances without data augmentation on the three datasets. 
Figure \ref{exp:data_augmentation} shows that the x-axis represents the number of times that we randomly shuffled the collected training set, and the y-axis represents the performance improvement percentages in a downstream task. 
We consistently observed that the downstream task performance improved as the number of feature shuffle operations increased. 
This indicates that shuffling-based augmentation can diversify training data, overcome sparsity, and boost the effectiveness of generated feature subsets.

\subsubsection{Robustness Check}
This experiment aims to answer: \textit{Is our method robust when different
ML models are used as a downstream task?} 
Feature selection methods depend on evaluating the currently selected feature subset at each step with a downstream predictive model to identify the optimal feature subset. 
In our design, we employed the Random Forest for evaluation due to its stability and robustness. 
However, more rigorously, a feature selection method should be able to generalize to diverse downstream models and the selected features can achieve robust performances over diverse downstream models. 
For this purpose, we tested all the best feature subsets selected by baselines and our method with 5 widely used downstream predictive models, including support vector machine (SVM), XGBoost (XGB), K-nearest neighborhood (KNN), and decision tree (DT) with the SVMguide3 dataset in terms of the F1 score. Table~\ref{exp:robustness_check} shows that, regardless of the downstream model used to evaluate the optimal feature subset generated by FSNS, the F1 score outperforms that of the baseline and the original feature set. Therefore, our method exhibits downstream model-agnostic robustness and reliability. The underlying driver is that we collect data corresponding to downstream models to train our model, formulate specific optimization strategies  and achieve robust optimization search.


\begin{table}[t]
    \centering
    \caption{Robustness check results. We compare model performance under different downstream models on the SVMGuide3 dataset.}
    \label{exp:robustness_check}
    \resizebox{70mm}{!}{
    \begin{tabular}{cccccc}
   \toprule\toprule
        & DT & KNN & SVM & XGB & RF\\
    \midrule
    Original & 76.38 & 73.97 & 75.11 & 81.31 & 77.81 \\
    K-Best & 77.85 & 72.91 & 75.08 & 76.12 & 76.54 \\
    mRMR & 76.08 & 73.96 & 75.08 & 75.49 & 76.54 \\
    LASSO & 76.41 & 73.45 & 75.11 & 74.72 & 75.29 \\
    RFE & 75.93 & 74.40 & 76.07 & 78.50 & 79.43 \\
    GFS & 77.22 & 75.01 & 73.87 & 78.69 & 76.95\\
    MCDM & 77.90 & 74.87 & 75.11 & 80.89 & 80.25\\
    LASSONet & 76.83 & 69.19 & 74.09 & 79.34 & 74.87 \\
    SARLFS & 74.39 & 72.30 & 70.52 & 78.65 & 82.48 \\
    MARLFS & 76.09 & 72.36 & 76.38 & 76.62 & 78.32 \\
    \midrule
    FSNS & \textbf{78.37} & \textbf{76.18} & \textbf{77.40} & \textbf{82.26} & \textbf{83.51}\\
    \bottomrule\bottomrule
    \end{tabular}}
\end{table}



\subsubsection{Examining the Scalability with Respect to Feature Size}
\begin{figure}[t]
    \centering
    \subfigure[Training time]{
    \begin{minipage}[ht]{0.23\linewidth}
    \centering
    \includegraphics[width=1.5in]{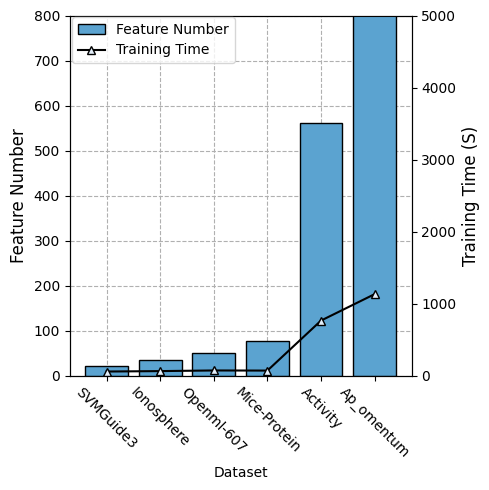}
    \end{minipage}
    }
    \centering
    \subfigure[Data collection time]{
    \begin{minipage}[ht]{0.23\linewidth}
    \centering
    \includegraphics[width=1.5in]{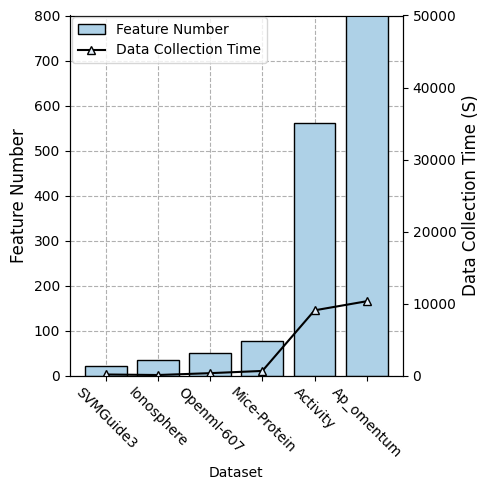}
    \end{minipage}
    }
    \centering
    \subfigure[Search time]{
    \begin{minipage}[ht]{0.23\linewidth}
    \centering
    \includegraphics[width=1.5in]{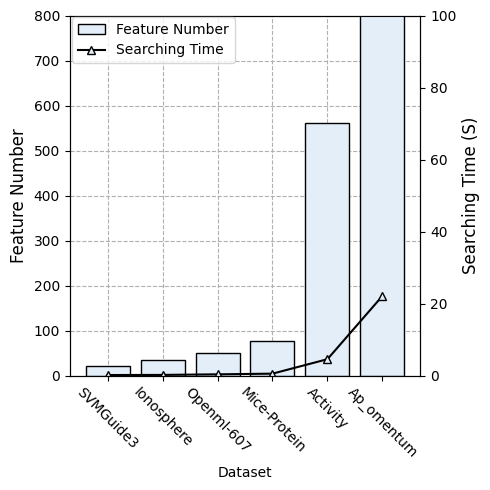}
    \end{minipage}
    }
    \centering
    \subfigure[Parameter size]{
    \begin{minipage}[ht]{0.23\linewidth}
    \centering
    \includegraphics[width=1.5in]{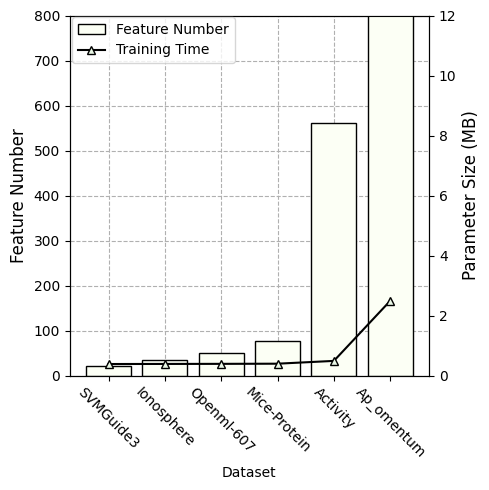}
    \end{minipage}
    }
    \caption{Results of time and space complexity. We study the training time, data collection time, search time, and parameter size of FSNS as  the number of features increase.} \label{exp:complexity}
    \vspace{-0.3cm}
\end{figure}

This experiment aims to answer: \textit{Is the time and space complexity of our model relatively stable as the number of features increases?} In other words, can our method efficiently handle large feature sets while maintaining acceptable running time and manageable parameter sizes? 
We selected 7 datasets with feature sizes ranging from 21 (the svmguide3 dataset) to 10,936 (ap\_omentum\_ovary dataset, referred to as ap\_10936 in Figure \ref{exp:complexity} to evaluate the training time, data collection time, search time, and needed parameter size with respect to feature number. Figure \ref{exp:complexity} shows that, as the feature size increases from 21 of svmguide3 to 10,936 of ap\_omentum\_ovary (470 times that of 21), The time spent on training, data collection, and searching increased by 20, 74, and 169 times, respectively. The parameter sizes increase 8 times compared to their original parameter size with 21 features. This is because, during the training phase, tokens are embedded into a low-dimensional space, making the time complexity and space complexity robust to the increase in feature dimensions. In the data collection phase, the time is primarily influenced by the computation of downstream model accuracy. However, the overall growth rate is acceptable, and our provided unsupervised collection method addresses this issue. Gradient-steered optimization search is generally time-efficient, even when the feature dimension is 10,936, a single search takes approximately 20 seconds, which has a minimal impact on the overall time complexity of the model.
Therefore, our method demonstrates promising scalability. 

\begin{figure}[htp]
    \centering
    \subfigure[Performance]{
    \begin{minipage}[ht]{0.4\linewidth}
    \centering
    \includegraphics[width=2in]{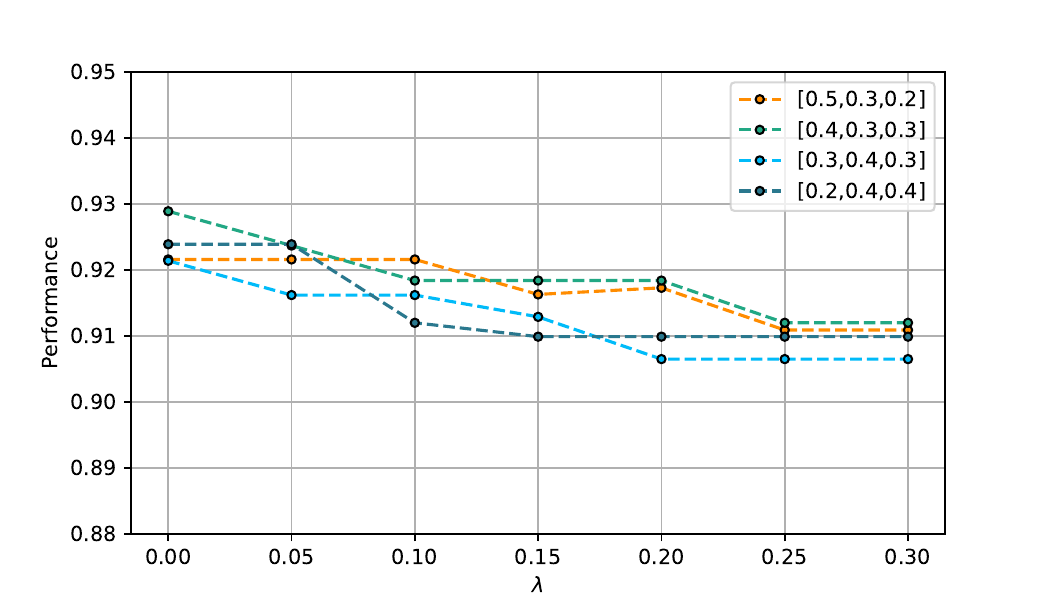}
    \end{minipage}
    }
    \centering
    \subfigure[Redundancy]{
    \begin{minipage}[ht]{0.4\linewidth}
    \centering
    \includegraphics[width=2in]{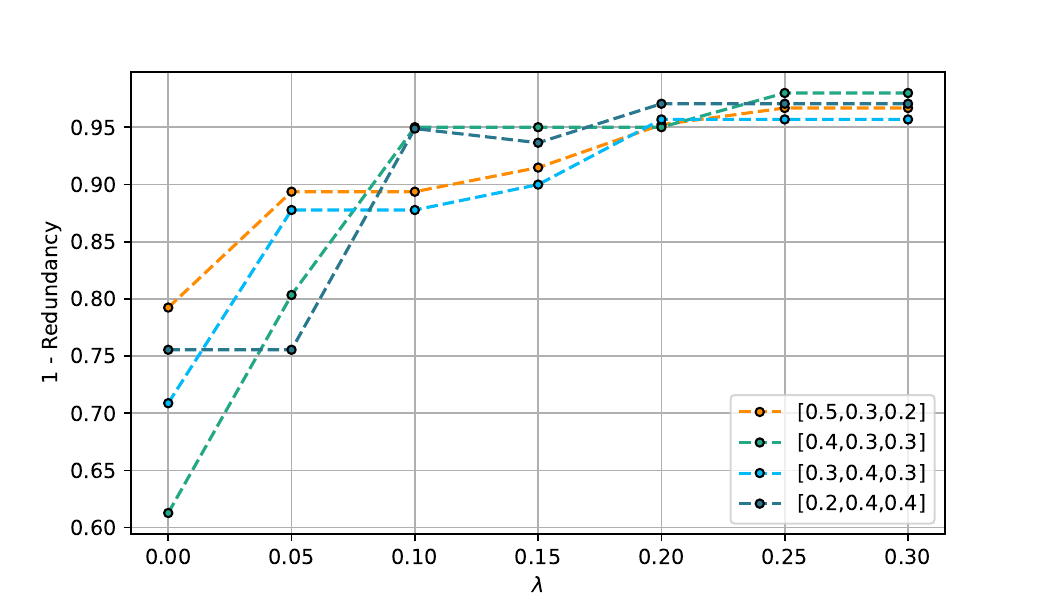}
    \end{minipage}
    }
    \caption{The hyparameter sensitivity analysis on SpamBase. The different colored lines denote different loss weight configurations, represented by [$\alpha,\beta, \delta$]. The horizontal axis represents the values of $\lambda$, while the vertical axis corresponds to performance and 1 - redundancy, respectively.}
    \label{exp:parameter}
    \vspace{-0.7cm}
\end{figure}

\subsubsection{A Study of the Hyperparameter Sensitivity} This experiment aims to answer: \textit{Is the performance of FSNS sensitive to values of the trade-off parameters?} We investigated the key hyperparameters, including the loss weight parameters $\alpha, \beta, \delta$, and the search trade-off parameter between performance and redundancy $\lambda$. 
We compared the performance and redundancy achieved by FSNS under different parameter configurations using the SpamBase dataset. The results are shown in Figure \ref{exp:parameter}, where the different colored lines denote different loss weight configurations (represented as [$\alpha, \beta, \delta$]) and the horizontal axis denotes different $\lambda$ values. We can observe that different configurations of $\alpha, \beta$, and $\delta$ slightly affect the results with the same $\lambda$ value. This suggests that during the training phase, FSNS is robust to the hyperparameters and can effectively model the data distribution under various weight configurations. $\lambda$ is the trade-off parameter between performance and redundancy during the search phase. A higher $\lambda$ value makes the model more inclined to find feature subsets with lower redundancy. These findings provide insights into how the hyperparameter impacts the results of FSNS and how to choose the values of the hyperparameters to balance the performance and redundancy.

\subsubsection{A study of the Traceability and Interpretability}
\begin{figure}[htp]
    \centering
    \subfigure[Original feature space]{
    \begin{minipage}[ht]{0.4\linewidth}
    \centering
    \includegraphics[width=2in]{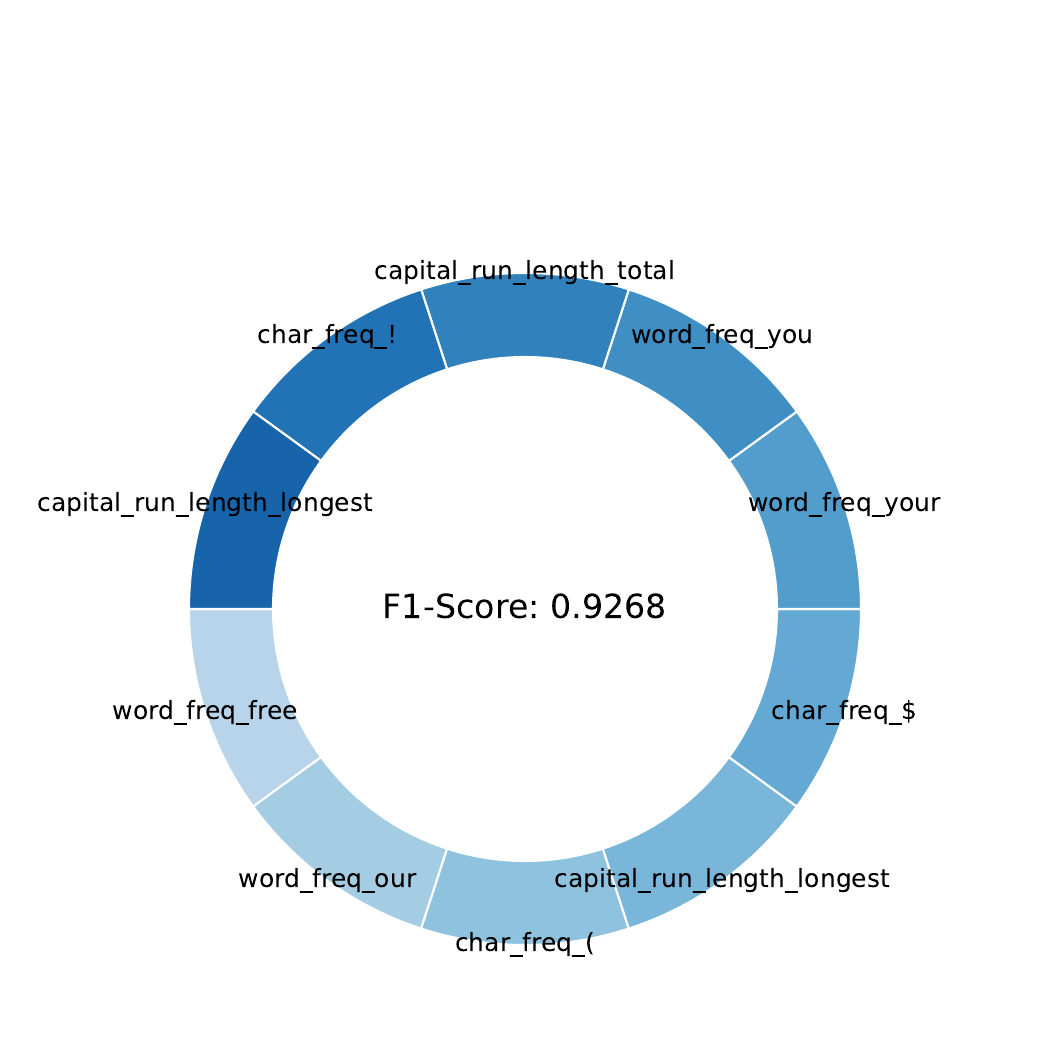}
    \end{minipage}
    }
    \centering
    \subfigure[FSNS selected feature space]{
    \begin{minipage}[ht]{0.4\linewidth}
    \centering
    \includegraphics[width=2in]{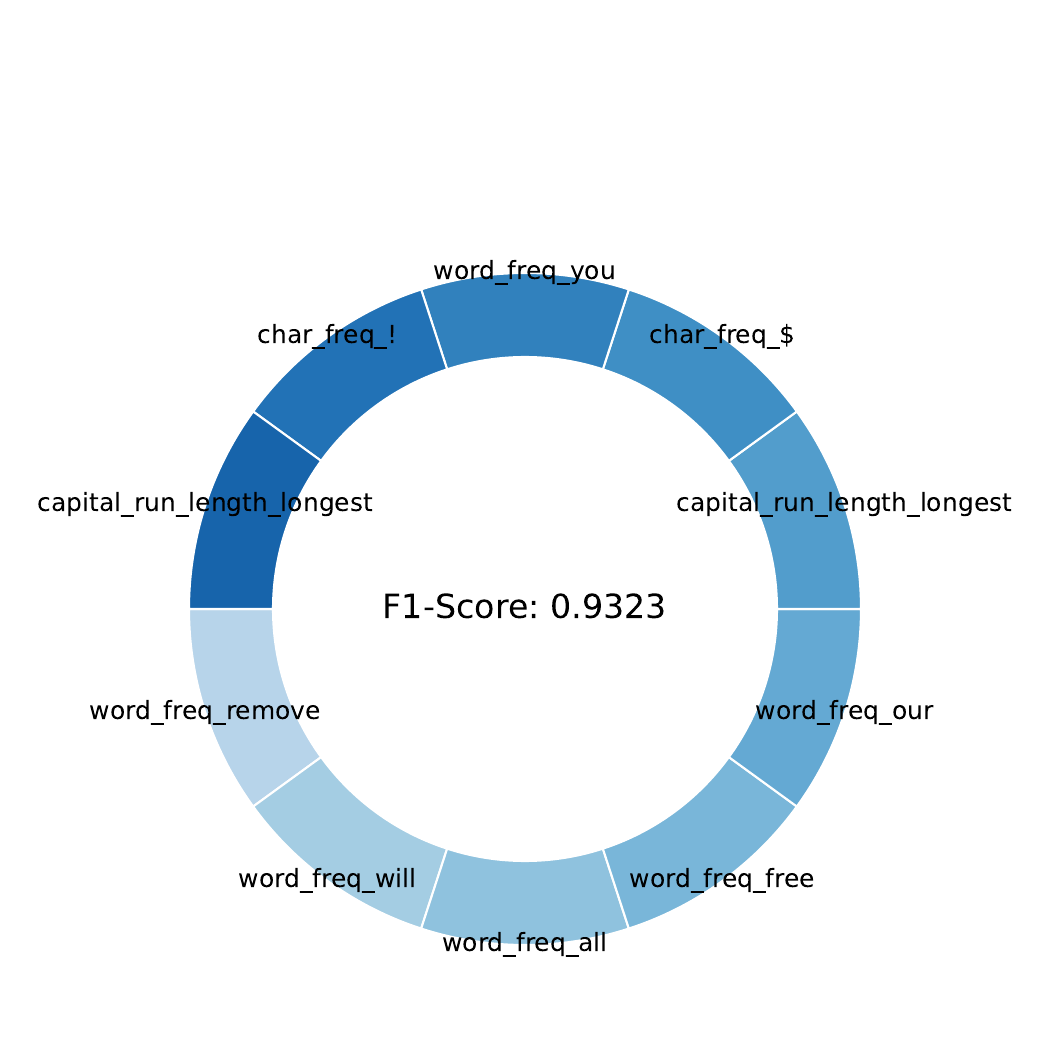}
    \end{minipage}
    }
    \caption{The top 10 important features identified in the original feature space and the FSNS selected feature space using the mutual information between features and label.}
    \label{exp:inter}
    \vspace{-0.3cm}
\end{figure}

This experiment aims to answer: \textit{Is the constructed feature space traceable and interpretable?} According to the mutual information between features and label, we report in Figure \ref{exp:inter} the top 10 essential features in the original and FSNS selected feature spaces respectively, with darker colors indicating more important features. We can observe that the valuable features (e.g., "capital\_run\_length\_longest") are retained by FSNS, and the redundant features such as (e.g., "word\_freq\_you") are eliminated. The underlying driver is that long sequences of uppercase letters are often used in spam emails to grab attention and emphasize urgency. But the word "you" serves a general communicative function. It is hard to distinguish spam emails according to the frequency of "you". In addition, new features (e.g., "word\_freq\_all") emerge among the top 10 features. This can be explained as “all” is frequently used to make broad promises or sweeping claims in spam emails. This result suggests that FSNS can explicitly trace the features through their names and interpret them through the semantic descriptions.

\subsubsection{Case Explanation: Our Method Exhibits Noise
Resistance.}
\begin{figure}[t]
	\centering
	\subfigure[Openml\_607]{\label{exp:case:607}\includegraphics[width=0.4\textwidth]{{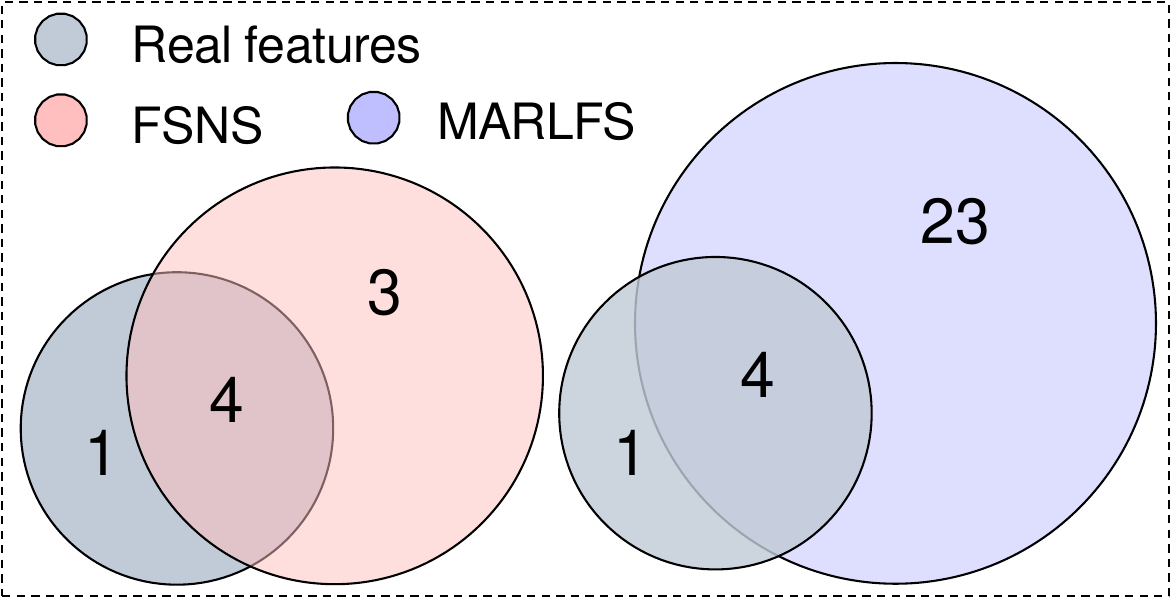}}}
	\subfigure[Openml\_618]{\label{exp:case:618}\includegraphics[width=0.4\textwidth]{{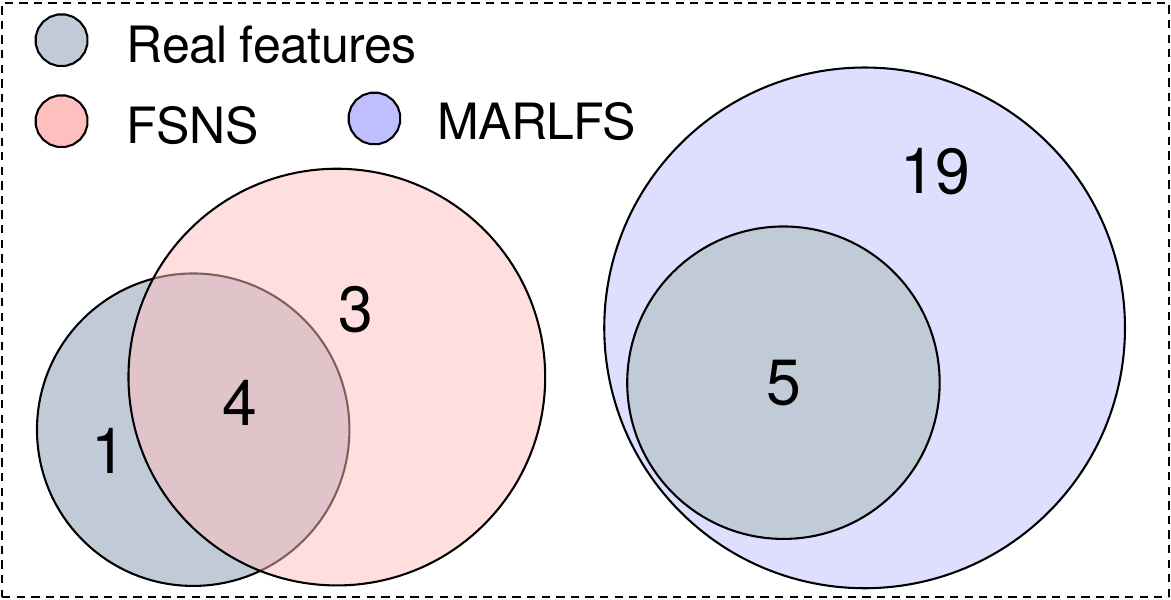}}}
	\caption{Case study in terms of our method and MARLFS based on the two OpenML datasets. Each dataset consists of 5 real features and 45 noise features. \textbf{FSNS \& MARLFS:} The number of feature subsets that our approach and RL-based method select from the OpenML dataset, respectively.}
	\label{exp:case_study}
 \vspace{-0.3cm}
\end{figure}

This experiment aims to answer: \textit{Can our model resist noise and recognize high-quality features?} 
We compared our method with an automated discrete choice model: MARLFS \cite{liu2019automating} on openml\_607 and openml\_618. Both datasets consist of 5 real features and 45 noise features. 
Figure \ref{exp:case_study} shows: 1) our method selected 4 informative features but only 3 noise features from on openml\_607; 2) MARLFS selects 4 informative features but 23 noise features. Similarly, on the data set of openml\_618, our method selects 4 informative features and MARLFS selects 5 informative features. But MARLFS chose far more noise features than our method. 
Our method demonstrates the ability to effectively denoise features and identify high-quality features.

\section{Related Work}

\subsection{Feature Selection}
Feature selection is crucial in improving the performance of downstream models \cite{peng2005feature,wang2024knockoff,ying2024feature,ying2023self,ying2024topology,hu2024reinforcement,ying2024unsupervised,gong2024evolutionary}. Various classical works are widely used. For instance, statistical methods including the Pearson correlation test \cite{meng1992comparing}, the student’s t-test \cite{owen1965power}, the Wilks’ lambda test \cite{el2007feature}, complementary and consensus learning~\cite{huang2023imufs} and the ANOVA F-test \cite{st1989analysis}. These methods typically assess the importance of each feature from a statistical perspective, where higher scores indicate greater importance of the corresponding feature. In addition, the least absolute shrinkage and selection operator (LASSO) \cite{tibshirani1996regression} adds L1-regularization to simple linear regression, which can shrink the coefficients of unimportant variables to zero, thereby achieving feature selection. Decision tree algorithm has also been applied to feature selection \cite{sun2017attribute}, evaluating the importance of features by information entropy.


In recent years, feature selection has been regarded as a discrete searching problem. Optimization algorithms such as metaheuristic algorithms \cite{sayed2019feature,al2019survey,zorarpaci2016hybrid} and reinforcement learning \cite{wang2022group,xiao2023traceable,xiao2024traceable} show promising potential in this filed. Some metaheuristic algorithms including particle swarm optimization (PSO) \cite{ajibade2020hybrid}, genetic algorithm (GA) \cite{oh2004hybrid}, grasshopper optimization algorithm (GOA) \cite{al2019survey}, Crow search algorithm (CSA) \cite{sayed2019feature}, bee colony optimization (BCO) \cite{zorarpaci2016hybrid}, and grey wolf optimizer (GWO) \cite{al2019binary} have been applied to feature selection. These methods iteratively generate and evaluate new solutions. Reinforcement learning-based methods \cite{liu2019automating,liu2023interactive,liu2021efficient} create agents to take actions in deciding the selection and deselection of the features. Each action yields a reward to assess its effectiveness, aiming to improve subsequent decisions on action. 

Due to its powerful feature representation capability and generalizability, deep learning plays significant roles in various fields~\cite{xiao2023beyond,wang2024reinforcement}. Some existing studies try to incorporate deep learning into feature selection, achieving competitive results. Li et al. \cite{li2016deep} develop a deep feature selection (DFS) model. Using elastic net, they introduce a sparse one-to-one linear layer between the input layer and the first hidden layer of a multi-layer perceptron. Then, the important features are selected according to their weights in the input layer. Also in \cite{shrikumar2017learning}, a method called Deep Learning Important Features (DeepLIFT) is proposed to score features using a back propagation-like algorithm. Nezhad et al. \cite{nezhad2016safs,nezhad2018predictive} employ autoencoder to obtain high-level feature representations. Then, the features are selected by a classical feature selection method and applied to a downstream model. The number of nodes in the autoendoer hidden layers, is adjusted to continuously optimize the structure of the network based on the downstream model results.

\subsection{Deep Sequential Generative AI \& LLM}
Deep generative models \cite{goodfellow2014generative,arjovsky2017wasserstein,kingma2013auto} have demonstrated immense potential in representing complex distributions in high-dimensional spaces. This property enables them to still exhibit excellent performance when dealing with challenging sequential data such as text \cite{kurup2021evolution}, music \cite{briot2021artificial}, video streams \cite{huang2017analysis}, etc. Particularly,  Large Language Models (LLMs) including GPT \cite{radford2018improving}, BERT \cite{devlin2018bert}, T5 \cite{raffel2020exploring}, have greatly revolutionized the development of AI in recent years. These models are extensively trained on massive datasets to understand and generate text that closely mimics human language. Since complex human knowledge can be modeled and generated by such deep sequential generative models, following the similar spirit, we attempt to serialize feature subsets to learn feature knowledge in continuous space, thereby conducting optimal feature subset selection.

\section{Conclusion}
In this paper, we regarded feature selection as a sequence of selection decision tokens and represent feature knowledge in a token symbolic representation. 
We solved feature selection from a neuro-symbolic learning perspective to leverage  generative AI to learn feature knowledge from symbolic token sequences and generate feature selection decisions. 
We implemented a four-step framework to map feature subset into an embedding space for optimizing feature selection: 1) using reinforcement learning to collect feature subset-accuracy data pairs; 2) constructing a feature subset embedding space using a variational transformer-based encoder-decoder-evaluator framework; 3) searching for the optimal feature subset embedding based on the gradient of the evaluator; 4) reconstructing the feature subset with the decoder. 
Our research findings indicate that: 1) reinforcement learning data collector is a tool for automated training data collection, enhancing the diversity, and scale of training data; 2) the encoder-decoder-evaluator framework effectively constructs the feature subset embedding space and improves effectiveness and task-agnosticism; and 3) the gradient-based search strategy generates gradient and direction information to effectively steer the gradient ascent-based search and identify the optimal feature subset.

\section*{Acknowledgment}
This research was partially supported by the National Science Foundation (NSF) via the grant numbers: 2426340, 2416727, 2421864, 2421865, 2421803, and National academy of engineering Grainger Foundation Frontiers of Engineering Grants.

\bibliographystyle{ACM-Reference-Format}
\bibliography{refer}

\vfill

\end{document}